\documentclass[11pt, a4paper, logo, copyright]{googledeepmind}

\pdftrailerid{redacted}

\makeatletter
\renewcommand\bibentry[1]{\nocite{#1}{\frenchspacing\@nameuse{BR@r@#1\@extra@b@citeb}}}
\makeatother

\usepackage{kantlipsum, lipsum}
\usepackage{dsfont}
\usepackage{gdm-colors}

\usepackage[authoryear, sort&compress, round]{natbib}

\usepackage[utf8]{inputenc} 
\usepackage[T1]{fontenc}    
\usepackage{hyperref}       
\usepackage{url}            
\usepackage{booktabs}       
\usepackage{nicefrac}       
\usepackage{microtype}      
\usepackage{amsmath}
\usepackage{graphicx}
\usepackage{multicol}
\usepackage[nameinlink]{cleveref}
\usepackage{bbm}
\usepackage{multirow}
\usepackage{subfig}
\usepackage{soul}
\usepackage{floatrow}
\usepackage{float}
\usepackage{wrapfig}
\usepackage{blindtext}
\usepackage{tablefootnote}
\usepackage{amsfonts}
\usepackage[flushleft]{threeparttable}
\usepackage{colortbl}
\usepackage{bbding}

\DeclareMathOperator*{\argmax}{arg\,max}

\usepackage{xspace}

\newcommand{\ours}{Gecko\xspace}
\newcommand{\dataset}{FRet}

\newfloatcommand{capbtabbox}{table}[][\FBwidth]

\newcommand{\draftonly}[1]{#1}
\newcommand{\eat}[1]{}

\newcommand{\draftcomment}[3]{\draftonly{\textcolor{#2}{{{[#1: #3]}}}}}

\newcommand{\jinhyuk}[1]{\draftcomment{Jinhyuk}{red}{#1}}

\crefformat{section}{\S#2#1#3}

\graphicspath{{figures/}}

\title{Gecko: Versatile Text Embeddings Distilled from Large Language Models}

\correspondingauthor{jinhyuklee@google.com}

\renewcommand{\today}{}

\author[*]{Jinhyuk Lee}
\author[*]{Zhuyun Dai}
\author[*]{Xiaoqi Ren}
\author[ ]{Blair Chen\hspace{-0.4ex}}
\author[ ]{Daniel Cer\hspace{-0.4ex}}
\author[ ]{Jeremy R. Cole\hspace{-0.4ex}}
\author[ ]{Kai Hui\hspace{-0.4ex}}
\author[ ]{Michael Boratko\hspace{-0.4ex}}
\author[ ]{Rajvi Kapadia\hspace{-0.4ex}}
\author[ ]{Wen Ding\hspace{-0.4ex}}
\author[ ]{Yi Luan\hspace{-0.4ex}}
\author[ ]{Sai Meher Karthik Duddu\hspace{-0.4ex}}
\author[ ]{Gustavo Hernandez Abrego\hspace{-0.4ex}}
\author[ ]{Weiqiang Shi\hspace{-0.4ex}}
\author[ ]{Nithi Gupta\hspace{-0.4ex}}
\author[ ]{Aditya Kusupati\hspace{-0.4ex}}
\author[ ]{Prateek Jain\hspace{-0.4ex}}
\author[ ]{Siddhartha Reddy Jonnalagadda\hspace{-0.4ex}}
\author[ ]{Ming-Wei Chang\hspace{-0.4ex}}
\author[ ]{Iftekhar Naim}

\affil[*]{Equal contributions}

\begin{abstract}
We present \ours{}, a compact and versatile text embedding model.
\ours{} achieves strong retrieval performance by leveraging a key idea: distilling knowledge from large language models (LLMs) into a retriever.
Our two-step distillation process begins with generating diverse, synthetic paired data using an LLM.
Next, we further refine the data quality by retrieving a set of candidate passages for each query, and relabeling the positive and hard negative passages using the same LLM. 
The effectiveness of our approach is demonstrated by the compactness of the Gecko. On the Massive Text Embedding Benchmark (MTEB), Gecko with 256 embedding dimensions outperforms all existing entries with 768 embedding size. Gecko with 768 embedding dimensions achieves an average score of 66.31, competing with 7x larger models and 5x higher dimensional embeddings.
\end{abstract}

\begin{document}

\maketitle

\section{Introduction}


Text embedding models represent natural language as dense vectors, positioning semantically similar text near each other within the embedding space~\citep{le2014distributed,reimers2019sentence,gao2021simcse}. These embeddings are commonly used for a wide range of downstream tasks including document retrieval, sentence similarity, classification, and clustering~\citep{muennighoff2023mteb}. Instead of building separate embedding models for each downstream task, recent efforts seek to create a single embedding model supporting many tasks.

The recent development of general-purpose text embedding models presents a challenge: these models require large amounts of training data to comprehensively cover desired domains and skills. Recent embedding efforts have focused on using extensive collections of training examples~\citep{wang2022text,li2023towards}.
Large language models (LLMs) offer a powerful alternative, as they contain vast knowledge across various domains and are known to be exceptional few-shot learners~\citep{brown2020gpt, anil2023palm}. Recent work demonstrates the effectiveness of using LLMs for synthetic data generation, but the focus has primarily been on augmenting existing human-labeled data or improving performance in specific domains~\citep{dai2022promptagator,jeronymo2023inpars}.
It motivates us to re-examine: to what extent can we leverage LLMs directly to improve text embedding models?

In this work, we present \ours{}, a highly versatile yet efficient embedding model, powered by the vast world knowledge of LLMs.
Our approach leverages insights from knowledge distillation to create a two-step LLM-powered embedding model.
Starting with a large corpus of (unlabeled) passages, we use a few-shot prompted LLM to generate a relevant task and query for each passage,
similar to \citet{dai2022promptagator} and \citet{wang2023improving}.
We then embed the concatenated task and query using a pretrained embedding model to obtain nearest neighbor passages, use an LLM to rerank the passages, and obtain positive and negative passages based on the LLM scores.
The reranking step is key to enhance the quality as we discover that the best passage to answer the generated query often differs from the original source passage.
We show that using our LLM-based dataset, \dataset{}, alone can lead to significantly improvement, setting a strong baseline as a zero-shot embedding model on MTEB.

By combining this LLM-generated and LLM-ranked data with human-annotated data, our model, \ours{}-1B with 768-dimensional embeddings, achieves the best performance on the popular MTEB benchmark~\citep{muennighoff2023mteb} among the models with compatible embedding dimensions and model sizes.
Moreover, \ours{} often outperforms other systems that use either larger base models (7B) or higher dimensional embeddings (1k to 4k).
\begin{figure}[t]
	\centering
	\includegraphics[width=0.87\columnwidth]{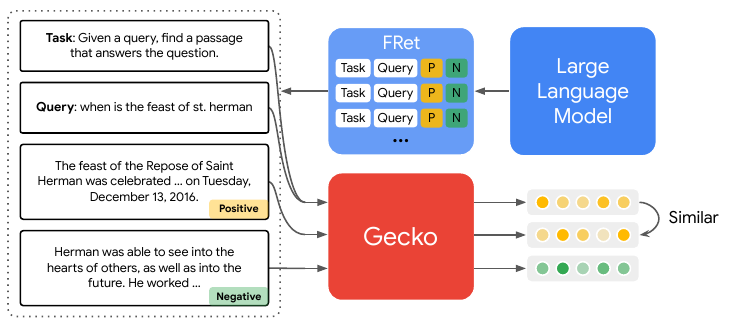}
	\caption{
	Overview of \ours{}.
	\ours{} is a versatile text embedding model trained on a variety of tasks including document retrieval, semantic similarity, and classification.
	To train \ours{}, we utilize \dataset{} where queries are generated from LLMs, and their positive and negative passages are mined by LLMs.
	}
	\label{fig:gecko_overview}
\end{figure}

\section{Related Work}
\paragraph{Text Embedding Models}





Text embeddings convert textual inputs into uniform-sized vectors, supporting downstream tasks such as semantic similarity, information retrieval, clustering, and classification. Recent models, including SBERT~\citep{reimers2019sentence}, Universal Sentence Encoder~\citep{cer2018universal}, and Sentence T5~\citep{ni2022sentence}, attempt to provide general purpose embeddings suitable for various NLP tasks. Despite attempting to be general-purpose, studies indicate that these embedding models struggle to generalize across tasks and domains,
motivating the creation of unified models trained across diverse tasks~\citep{su2022one,asai2022task} and benchmarks such as MTEB~\citep{muennighoff2023mteb} focused on novel task and domain generalization.
Inspired by these prior works, we develop a versatile embedding model by creating the LLM-generated \dataset{} dataset from a large and diverse corpus encompassing a wide variety of task types. 

\paragraph{Contrastive Learning}
One of the critical components of contrastive learning is to find proper negative examples for a query~\citep{gao2021simcse,Karpukhin2020DensePR,lee2021learning}.
For example, \citet{xiong2020approximate} proposed to select hard negatives from a large corpus using an asynchronously-updated approximate nearest neighbor index.
Other previous work has denoised the hard negatives based on confidence scores~\citep{qu2021rocketqa,ren2021rocketqav2} or distilled knowledge from cross-attention rerankers into the dual-encoders ~\citep{izacard2021distilling,santhanam2022colbertv2,sachan2023questions}.
In our work, using LLMs, we study the effect of mining better positive examples for a query while finding useful hard negatives as well.
While similar in spirit to previous distillation approaches, using this hard selection of positive and negative passages aligns well with the format of existing human-annotated training data, allowing us to train on both.

\paragraph{Synthetic Data Generation}
When applying text embedding models to new tasks and domains, we often want to have relevant queries and labels for these target domains, but they are often unavailable or prohibitively expensive to collect.
To address this issue, several works~\citep{dai2022promptagator,bonifacio2022inpars,jeronymo2023inpars,khramtsova2024leveraging} propose a few-shot prompted query generation approach.
They generate synthetic queries by few-shot prompting LLMs to create a domain-specific training dataset, which has been shown to be very successful on the zero-shot information retrieval benchmark~\citep{thakur2021beir}.
In contrast to generating domain-specific queries for domain adaptation, our work aims to distill more general-purpose knowledge of LLMs into a text embedding model, resulting in a versatile text embedding model that achieves strong performance on MTEB~\citep{muennighoff2023mteb}.

\paragraph{Retrieval with Instructions}
Previously, \citet{dai2022promptagator} demonstrated that there exist different intents for different retrieval tasks.
For instance, given a search query, users might want to find a similar query, or they might want to read a passage that directly answers the query. Recent work has explored implementing a retriever that changes the retrieval behavior for different intents.
\citet{asai2022task} and \citet{su2022one} introduce ``retrieval with instructions,'' where a dense retriever is trained to follow an instruction that was given along with the query.
\citet{wang2023improving} also explores how LLMs can generate synthetic task instructions and associated queries, but for more general-purpose text embeddings similar to ours.
They use a two-step prompt to encourage the diversity of the synthetic data: first prompting an LLM to come up with a task and then generating an example (query, positive passage, and negative passage) based on the task.
In our work, we also synthesize task-query pairs to increase the diversity of the synthetic data.
Unlike \citet{wang2023improving}, however, we generate synthetic task and query pairs from the web passages, basing our \dataset{} dataset on real user-facing content.
We also use LLMs to decide which web passages can be used as positive or negative targets for each generated query.

\section{Training Recipe for \ours{}}
\ours{} is based on a 1.2B parameter pre-trained transformer language model that undergoes two additional training stages: pre-finetuning and fine-tuning.
First, we extend the pre-finetuning recipe from previous work (\citealp{Ni2021LargeDE}; \cref{sec:prefinetuning}).
For fine-tuning, our main contribution is to create a novel fine-tuning dataset for a diverse set of downstream tasks via a two-step LLM distillation, which identifies both positive and hard negative passages for each generated query (\cref{sec:fret}).
We coin this dataset as \dataset{}, the \textbf{F}ew-shot Prompted \textbf{Ret}rieval dataset.
For the fine-tuning mixture, \dataset{} is combined with a diverse set of academic datasets formatted in a similar way: each with a task description, input query, positive passage, and negative passage (\cref{sec:mixture}).

\subsection{Pre-finetuning}\label{sec:prefinetuning}
Following the prior work~\citep{Ni2021LargeDE,neelakantan2022text,wang2022text}, our pre-finetuning procedure relies on self-supervised tasks over a large text corpus as described below.

\paragraph{Training Mixture} We use two pre-finetuning datasets.
First, we use the large-scale community QA dataset by \citet{Ni2021LargeDE}, which includes text pairs such as question-answer pairs from online forums and QA websites.
Next, we crawl a corpus of title-body text pairs from the Web, which can be found from almost every website as naturally occurring pairs.
Despite its simplicity, \citet{wang2022text} showed that these naturally occurring text pairs are useful for pre-finetuning embedding models.

\paragraph{Training Objective}
Pre-finetuning on a large amount of unsupervised text pairs has been shown to improve performance for smaller-scale dual encoders for various downstream tasks including document retrieval~\citep{lee2019latent,izacard2021contriever} and semantic similarity~\citep{gao2021simcse}.
The goal of the pre-finetuning stage is to expose the model to a large amount of textual diversity, which seems necessary for the compact text embedding models that we aim to train.

We begin with a pre-trained language model $\mathcal{M}$ where $\mathcal{M}$ outputs a series of contextualized token embeddings $\mathbf{W} \in \mathbb{R}^{n \times d}$ given a sequence of $n$ tokens and an embedding dimension of $d$.
Given a set of text pairs $\mathcal{D}_\text{pre} = \{(q_i, p_i)\}_{i=1}^N$ for pre-finetuning, 
we obtain the vector representations of $q_i$ and $p_i$ by taking the mean of $\mathbf{W}$ along the $n$ axis. We first prepend a dataset-specific task feature $t$ before each query, so each query is informed of which task is being optimized.
\begin{equation}\label{eq:text_rep}
\begin{split}
    \mathbf{q}_i &= \texttt{mean\_pool}_{\lvert t \rvert + \lvert q_i \rvert}\left[\mathcal{M}(t \oplus q_i) \in \mathbb{R}^{(\lvert t \rvert + \lvert q_i \rvert) \times d}\right] \in \mathbb{R}^d \\
    \mathbf{p}_i &= \texttt{mean\_pool}_{\lvert p_i \rvert } \left[\mathcal{M}(p_i) \in \mathbb{R}^{\lvert p_i \rvert \times d} \right] \in \mathbb{R}^d. \\
\end{split}
\end{equation}
For pre-finetuning, we use simple task features such as \textit{question answering} or \textit{search result} for $t$ depending on the dataset.
Then, for each mini-batch of size $B$, we optimize the contrastive learning objective with in-batch negatives:
\begin{equation}\label{eq:contrastive}
    \mathcal{L}_\text{pre} = \frac{1}{B}\sum_{i=1}^B \\ \left[-\log \frac{e^{\text{sim}{(\mathbf{q}_i, \mathbf{p}_i})/\tau}}{\sum_{j=1}^B e^{\text{sim}(\mathbf{q}_i, \mathbf{p}_j)/\tau}}\right].
\end{equation}
In this work, we use the cosine similarity for the similarity function, $\text{sim}(\mathbf{x}, \mathbf{y}) = \frac{\mathbf{x}^\top\mathbf{y}}{||\mathbf{x}|| \cdot ||\mathbf{y}||}$, with a temperature parameter $\tau$.
Note that we do not utilize hard negatives during pre-finetuning and utilize the maximum batch size that fits into the device.
This has been found to be effective for document retrieval tasks as observed in previous work~\citep{wang2022text,li2023towards}.

\subsection{\dataset{}: Two-Step LLM Distillation}\label{sec:fret}
\begin{figure}[t]
	\centering
	\includegraphics[width=1.0\columnwidth]{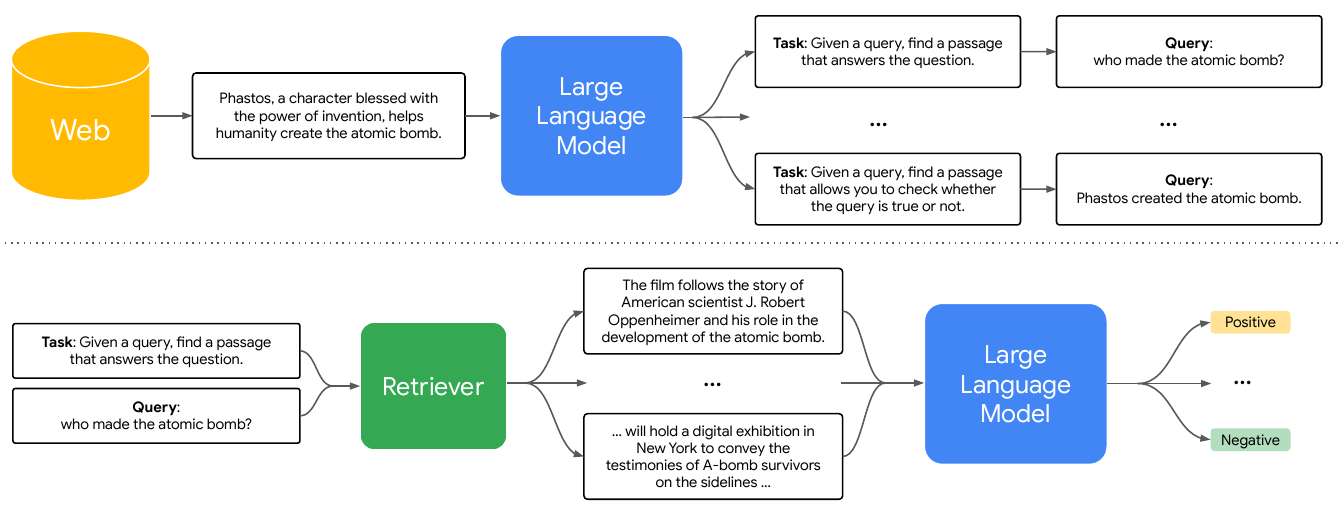}
	\caption{Overview of \dataset{}.
	Given a sampled passage from the web, \dataset{} first utilizes LLMs to generate a relevant task and a query for the passage (top).
	Then, each query and task is fed into a pre-trained embedding model to obtain nearest neighbor passages, which are then scored by the LLM to mine positive and negative passages (bottom).
	Note that the original web passage does not necessarily become a positive passage as LLMs can find a more relevant passage as shown above.
	}
	\label{fig:fret_overview}
\end{figure}

In this section, we introduce our two-stage approach that uses LLMs to generate \dataset{}. Traditional approaches for training embedding models often rely on large, manually labeled datasets. However, creating such datasets is time-consuming, expensive, and often results in undesirable biases and lack of diversity. In this work, we present a novel method for generating synthetic data for training multi-task text embedding models, leveraging the power of LLMs through a two-step distillation process.
The overall process of generating \dataset{} is illustrated in \Cref{fig:fret_overview}.



\paragraph{LLM-based Diverse Query Generation}
One of the challenges of using manually crafted queries is to ensure that the queries cover a diverse set of tasks and linguistic patterns.
With LLMs, these variables are relatively easy to control as we can design the prompt to specify the diversity.
In this work, we employ few-shot prompts to control the diversity of queries.
Our LLM is instructed to read a sampled web passage and generate both the task description and a relevant query for the task:
\begin{equation*}
   \text{LLM}(\mathbb{P}_\text{QG},  p_\text{seed}) \rightarrow (t, q)
\end{equation*}
where $p_\text{seed}$ is a passage drawn randomly from the web corpus $\mathcal{C}$ and $\mathbb{P}_\text{QG}$ is a fixed prompt.
The prompt, $\mathbb{P}_\text{QG}$, is identical for every example and consists of few-shot examples and instructions.  
The LLM generates a task description $t$, which describes the type of retrieval---for example, \textit{`Given a query, find a passage that has the answer to the query'} (question answering) or \textit{`Given a query, find a passage that allows you to check whether the query is true or not'} (fact checking)---and also a query $q$ that aligns with the task.
By sampling over such free-form task descriptions, we guide the LLM to produce a wide range of queries.
These pairs are later used to train our embedding models, teaching the models to associate a query and its corresponding instructions with the target passage.

The diversity of \dataset{} comes from two sources.
First, a web corpus inherently contains a variety of topics as well as styles of writing, such as blog posts, news, Wikipedia-like content, and forum posts.
Second, by adding many diverse task descriptions in the prompt, we encourage the LLM to generate more diverse task descriptions and therefore more diverse queries.
Similar to \citet{dai2022promptagator}, our method can be applied to any corpus of passages. Our method is different from approaches such as~\citet{wang2023improving}, where LLMs generate both synthetic queries and synthetic passages.

\paragraph{LLM-based Positive and Negative Mining}
Most models that utilize synthetic queries are trained with $(q, p_\text{seed})$ pairs, which assumes that $p_\text{seed}$ is a good positive target for $q$~\citep{dai2022promptagator,jeronymo2023inpars}.
While this is likely true in most cases, we hypothesize that there could be a more relevant passage than $p_\text{seed}$ somewhere in our corpus of web passages.
Essentially, in the previous section, we sampled $\operatorname{P}(t, q \mid p_\text{seed})$ from the LLM, but this does not guarantee that $p_\text{seed}$ maximizes $\operatorname{P}(p \mid q, t)$ over all the passages in the corpus.
This intuition is supported by our observation that generated queries often focus on a particular aspect of a relatively long passage.
Hence, we propose a method that leverages LLMs to discover more relevant positive passages along with a good hard negative for the generated query.

In particular, we use an existing embedding model\footnote{In this work, we train an initial embedding model with $(q, p_\text{seed})$ pairs, treating in-batch passages as random negatives.} to retrieve top $N$ neighbors $P=\{p^{(1)}, \dots, p^{(N)}\}$ from the corpus given a generated query $q$.
We then employ the same LLM used for the query generation to rank these retrieved passages based on their relevance to the query. 
Specifically, we use two well-known few-shot prompted LLM ranking functions: query likelihood and relevance classification.
Query likelihood uses an LLM to measure the log-likelihood of a generated query $q$ given a passage $p$, i.e., $\text{QL}(q, p)=\text{LLM}(q \mid p, \mathbb{P}_\text{QL})$~\citep{Sachanql}. Herein, $\mathbb{P}_\text{QL}$ is a prompt containing an instruction for judging query likelihood and several few-shot examples of relevant query and passage pairs~\citep{drozdov2023parade}.
Relevance classification~\citep{zhuang2023beyond} uses an LLM to measure the log-likelihood of a specific relevance label given the query $q$ and a passage $p$, i.e., $\text{RC}(q, p)=\text{LLM}(\text{label}\mid q, p, \mathbb{P}_\text{RC})$, where $\mathbb{P}_\text{RC}$ is a prompt with few-shot examples for grading the relevance of each query-passage pair. The prompts  $\mathbb{P}_\text{QL}$ and $ \mathbb{P}_\text{RC}$ are identical for every example.
Our pilot study demonstrated that each prompting method (i.e. QL and RC) excels in different tasks, so we ensemble the rankings from two different prompting results with the standard Reciprocal Rank Fusion (RRF) approach~\citep{cormack2009reciprocal}, obtaining a ranking function $R(q, p)$. As shown in \Cref{sec:apdx_ensemble}, the ensembling greatly improves the  robustness of our model across diverse tasks.


Given the scores from LLMs after ensembling, we index the set of passages $P$ according to their ranking, i.e. $P = \{p_1, \ldots, p_N\}$ where if $i < j$, $R(q, p_i) \ge R(q, p_j)$. We then choose a new positive target: 
\begin{equation*}
p^+ = \argmax_{p\in P} R(q, p) = p_1
\end{equation*}
Importantly, $p^+$ can be different from $p_\text{seed}$ and conveys an approximation to the global preference of the LLM over the entire corpus.
\Cref{tab:fret_examples} lists examples where the $p^+$ differs from $p_\text{seed}$, demonstrating that the pair ($q, p_\text{seed}$) can be sub-optimal and there can be more relevant passages for $q$ globally.
We find that the relabeling of the positive passage (i.e., $p^+ \neq p_\text{seed}$) happens for about 15\% in our dataset.

Similarly, the LLM scores can also be used to select hard negative passages. One straightforward option is to select the lowest scoring negative, i.e. $p^- = p_{N}$. Another is to sample from the remaining nearest neighbors, i.e. $p^- \sim P\setminus\{p^+\}$. We explore both options in \Cref{sec:analysis}.
Combining all of our generation results along with the positive and negative mining, we create the \dataset{} dataset, comprised of 6.6M examples, each containing a task, a query, a positive passage, and a negative passage.

\subsection{Unified Fine-tuning Mixture}\label{sec:mixture}
We combine \dataset{} with other academic training datasets in the same format: task description, input query, positive passage (or target), and negative passage (or distractor), creating a novel fine-tuning mixture.
We then train our embedding model, \ours{}, using this mixture with a standard loss function.


\paragraph{Academic Data}
In addition to \dataset{}, we use the following academic training datasets: Natural Questions~\citep{kwiatkowski2019natural}, HotpotQA~\citep{yang2018hotpotqa}, FEVER~\citep{thorne2018fever}, MedMCQA~\citep{pal2022medmcqa}, SNLI~\citep{bowman2015large}, MNLI~\citep{williams2018broad}, and several classification datasets from Huggingface.
For the multilingual model, we add training sets from MIRACL~\citep{zhang2023miracl}.
All datasets are pre-processed to have a unified encoding format (\Cref{sec:apdx_format}), containing a task description, a query, a positive passage, and a negative passage.

\paragraph{Classification Data for Contrastive Learning}
We aim to seamlessly incorporate the classification training sets into our contrastive learning objective without any performance degradation on other tasks such as document retrieval.
Specifically, given a classification input text $x$ with a label $y \in \mathcal{Y}$, we pair each input $x$ with another input $x^+$, which shares the same label $y$ and then use $x^+$ as a positive target for $x$.
At the same time, we randomly select a hard negative input $x^-$ which has any label other than $y$.
This approach is a simple version of the classification datasets pre-processed by \citet{su2022one} but avoids using any model-specific embeddings.
During our experiments, we found that each $x^+$ might overlap with other positive examples within the mini-batch, creating a false negative problem among the in-batch negatives.
Hence, we assign a unique ID to each triple ($x$, $x^+$, $x^-$) and append the same unique ID to $x$, $x^+$, and $x^-$.
This effectively makes the in-batch negatives trivial for the model to distinguish them, because if the unique ID does not match, then it is never the correct answer. Thus, the model focuses on differentiating $x^+$ and $x^-$ given $x$. 

\paragraph{Training Objective}
For fine-tuning, we are given a set of $M$ fine-tuning datasets (including \dataset{}) that are comprised of a query-specific task description, an input, a positive target, and a hard negative: $[\mathcal{D}^{(1)}, \dots, \mathcal{D}^{(M)}]$ where $\mathcal{D}^{(m)} = \{(t_i, q_i, p_i^+, p_i^-)\}_{i=1}^N$.
We obtain the vector representations $\mathbf{q}_i$, $\mathbf{p}_i^+$, and $\mathbf{p}_i^-$ similar to \cref{eq:text_rep} where $t_i$ is used for the input: $\mathbf{q}_i = \texttt{mean\_pool}[\mathcal{M}(t_i \oplus q_i)]$.

For fine-tuning we optimize the in-batch cross-entropy loss, where query $q_i$ should distinguish $p_i^+$ from the hard negative $p_i^-$, other passages in the batch $\{p_j^+\}_{j=1}^B$, and other queries in the batch $\{q_j\}_{j=1}^B \setminus \{q_i\}$. The use of other queries in the batch is also known as "same-tower negatives"~\citep{moiseev2023samtone}.
Given a mini-batch of size $B$, we optimize the following objective:
\begin{equation}\label{eq:loss_main}
    \mathcal{L}_\text{main} = \frac{1}{B}\sum_{i=1}^B \left[-\log \frac{e^{\text{sim}(\mathbf{q}_i, \mathbf{p}_i^+)/\tau}}{\sum_{j=1}^B \left(e^{\text{sim}(\mathbf{q}_i, \mathbf{p}_j^+)/\tau} + \mathbbm{1}_{[j \neq i]}e^{\text{sim}(\mathbf{q}_i, \mathbf{q}_j)/\tau}\right) + e^{\text{sim}(\mathbf{q}_i, \mathbf{p}_i^-)/\tau}}\right].
\end{equation}
For the same-tower negatives, we used the indicator variable $\mathbbm{1}_{[j\neq i]}$ to denote that we are iterating over $j$ except for the current target index $i$.
Intuitively, same-tower negatives are helpful for symmetric text embedding tasks such as measuring the semantic similarity of two sentences, because $\{\mathbf{q}_j\}_{j=1}^B$ shares the same modality with $\mathbf{q}_i$: in this case, both are queries.
Finally, to support multiple different dimensions of embeddings with a single model, we add the MRL loss~\citep{kusupati2022matryoshka}, which optimizes \cref{eq:loss_main} with sub-dimensions smaller than $d$.
In our experiments, we use two embedding dimensions $d=768$ and $d=256$ for \ours{}.
\begin{table}[t]
    \small
    \centering
	\caption{Results on MTEB. We categorize models into two groups based on their embedding dimension (Dim.) and the number of parameters (\# Params.). We report the average performance on seven different tasks: Classification (Class.), Clustering (Cluter.), Pair Classification (Pair.), Reranking (Rerank.), Retrieval, STS, and Summary. The last column shows the average performance across all 56 datasets from the seven tasks. In the last row, we show the performance of a zero-shot \ours{} model, solely trained on \dataset{} without any human-labeled data or MTEB in-domain training datasets. Please refer to \Cref{sec:full_mteb} for the result and the instruction per dataset.
	}\label{tab:mteb}
	\resizebox{1.0\linewidth}{!}{%
	\begin{tabular}{lcc|ccccccc|c}
		\toprule
		& Dim. & \# Params. & Class. & Cluster. & Pair. & Rerank. & Retrieval & STS & Summary & Avg. \\
		
		\midrule
		\MakeLowercase{GritLM-8x7B} & 4,096 & 56B & 78.53 & 50.14 & 84.97 & 59.80 & 55.09 & 83.26 & 29.82 & 65.66 \\
		e5-mistral-7b-instruct & 4,096 & 7B & 78.47 & 50.26 & \textbf{88.34} & 60.21 & 56.89 & \textbf{84.63} & \textbf{31.40} & 66.63 \\
		echo-mistral-7b-instruct & 4,096 & 7B & 77.43 & 46.32 & 87.34 & 58.14 & 55.52 & 82.56 & 30.73 & 64.69 \\
		\MakeLowercase{GritLM-7B} & 4,096 & 7B & \textbf{79.46} & \textbf{50.61} & 87.16 & \textbf{60.49} & \textbf{57.41} & 83.35 & 30.37 & \textbf{66.76} \\
		text-embedding-3-large (OpenAI) & 3,072 & n/a & 75.45 & 49.01 & 85.72 & 59.16 & 55.44 & 81.73 & 29.92 & 64.59\\
 		
		\midrule
        gtr-t5-xxl & 768 & 5B & 67.41 & 42.42 & 86.12 & 56.66 & 48.48 & 78.38 & 30.64 & 58.97 \\
        gtr-t5-xl & 768 & 1.2B & 67.11 & 41.51 & 86.13 & 55.97 & 47.96 & 77.80 & 30.21 & 58.42 \\
        instructor-xl & 768 & 1.5B & 73.12 & 44.74 & 86.62 & 57.29 & 49.26 & 83.06 & 32.32 & 61.79 \\
        text-embedding-3-large-256 (OpenAI) & 256 & n/a & 71.97 & 46.23 & 84.22 & 57.99 & 51.66 & 81.04 & 29.92 & 62.00 \\
        \rowcolor{dmblue50} \textbf{\MakeLowercase{\ours{}-1B-256}} & 256 & 1.2B & 78.99 & 45.07 & 87.25 & 57.78 & 52.44 & 84.93 & 32.36 & 64.37 \\
        \rowcolor{dmblue50} \MakeLowercase{\textbf{\ours{}-1B-768}} & 768 & 1.2B & \textbf{81.17} & \textbf{47.48} & \textbf{87.61} & \textbf{58.91} & \textbf{55.70} & \textbf{85.06} & \textbf{32.63} & \textbf{66.31} \\
        \rowcolor{dmblue50} \MakeLowercase{-- zero-shot} (\dataset{}-only) & 768 & 1.2B & 70.26 & 46.82 & 86.27 & 57.60 & 53.16 & 83.14 & 32.16 & 62.64\\

        \bottomrule
	\end{tabular}
	}
\end{table}
\thisfloatsetup{subfloatrowsep=none}
\begin{figure}[t]
\begin{floatrow}
\capbtabbox{%
\resizebox{0.96\linewidth}{!}{%
\begin{tabular}{lc}
\toprule
    & MIRACL (Avg.) \\

    \midrule
    \textit{Per-language models} \\
    \midrule
    BM25 & 38.5 \\
    mDPR & 41.8 \\
    BM25 + mDPR (hybrid) & 56.6 \\
    \midrule
    \textit{One model for all languages} \\
    \midrule
    mDPR (en) & 39.7 \\ 
    mContriever (en) & 37.8 \\
    mContriever & 52.7 \\
    SWIM-X & 46.4 \\
    mContriever-X & 55.4 \\
    text-embedding-3-large (OpenAI) & 54.9 \\
    \rowcolor{dmblue50} \textbf{gecko-multilingual-1b} & \textbf{56.2} \\
\bottomrule
\end{tabular}
}
}{%
\caption{Results on MIRACL. We report average nDCG@10 on multilingual retrieval tasks in 18 languages (ar, bn, en, es, fa, fi, fr, hi, id, ja, ko, ru, sw, te, th, zh, de, yo).
Each row shows the performance of a single multilingual retriever.
}\label{tab:miracl}
}
\capbtabbox{%
\resizebox{0.964\linewidth}{!}{%
\begin{tabular}{ll|cc}
	\toprule
	Positive ($p^+$) & Hard Negative ($p^-$) & BEIR & STS \\ 
	\midrule
	\multicolumn{4}{l}{\textit{MS-MARCO}} \\
	\midrule
	$p_\text{seed}$ & None & 49.87 & \textbf{79.38} \\ 
	$p_\text{seed}$ & $p \sim P \setminus \{p_\text{seed}\}$ & 50.31 & 78.17 \\ 
	$p_1$ & $p \sim P \setminus \{p_1\}$ & 52.03 & 78.96 \\ 
	$p_1$ & $p_{20}$ & \textbf{52.29} & 78.96 \\ 
	\midrule
	\textit{\dataset{}} \\
	\midrule
	$p_\text{seed}$ & None & 52.33 & 82.66 \\
	$p_\text{seed}$ & $p \sim P \setminus \{p_\text{seed}\}$ & 51.37 & 82.00 \\
	$p_\text{seed}$ & $p_{20}$ & 51.96 & 82.26 \\

	$p_1$ & None & 53.07 & 82.88 \\
	$p_1$ & $p \sim P \setminus \{p_1\}$ & 52.60 & 82.85 \\
	$p_1$ & $p_{20}$ & \textbf{53.39} & \textbf{83.14} \\
    \bottomrule
\end{tabular}
}
}{%
\caption{
With MS-MARCO and \dataset{}, we test different strategies of choosing positive and hard negative passages.
We train each model and report its performance on BEIR (nDCG@10) and STS (Spearman Correlation) performance. 
}\label{tab:label_test}
}
\end{floatrow}
\end{figure}
\section{Experiments}
We mainly evaluate \ours{} on the Massive Text Embedding Benchmark (MTEB), which contains 56 datasets on retrieval, semantic textual similarity (STS), clustering, classification, pair classification, reranking, and summarization.
We analyze how each component of \ours{} and \dataset{} contribute to the performance, providing insights on building heterogeneous text embedding models.

\subsection{Main Results}
\Cref{tab:mteb} summarizes the performance of \ours{} and other baselines on MTEB.
For baselines, we report the performance of text embedding models whose recipes are fully (or partly) available.
\ours{} significantly surpasses all similarly-sized baselines (<= 1k embedding dimensions, <= 5B parameters) on every text embedding task in the MTEB benchmark. 
\ours{}-1b-256 demonstrates superior quality compared to text-embedding-3-large-256 (OpenAI; \citealt{neelakantan2022text}), GTR~\citep{Ni2021LargeDE}, and Instructor~\citep{su2022one}.  \ours{}-1b-768 often matches or exceeds the performance of even larger models, including text-embedding-3-large (OpenAI), E5-mistral~\citep{wang2023improving}, GRit~\citep{muennighoff2024generative}, and Echo embeddings~\citep{springer2024repetition}. Notably, these models all use 3-4k dimensional embeddings and exceed 7B parameters.
We observe that \ours{} is particularly good at balancing retrieval and STS performance, and sets a new state-of-the-art on classification, STS, and summary.
Surprisingly, the performance of \ours{} trained solely on \dataset{}, which makes MTEB a pure zero-shot benchmark, shows strong performance compared to other baselines.

\subsection{Multilingual Retrieval Results}
\Cref{tab:miracl} summarizes the performance of \ours{} and other baselines on MTEB.
We train a multilingual version of \ours{} with multilingual language models~\citep{xue2021mt5,team2023gemini} with the same recipe as \ours{}, but add the MIRACL training dataset in the mixture.
Note that \dataset{} is provided only in English and the main difference of gecko-multilingual-1b with others is the use of \dataset{} in its training set.
We find that while we only generated English-only dataset from LLMs, this translates well to other multilingual tasks achieving superior performance compared to others.

\begin{table}[t]
    \small
    \centering
	\caption{
	Does the diversity of \dataset{} matter when training versatile embedding models?
	We test different subsets of \dataset{} for training and report their performance on MTEB.
	From the four most frequent tasks in \dataset{} (e.g., \dataset{}-question-answering), we sample 300k training examples.
	For \dataset{}-all-tasks, we sample 75k training examples from each task to form 300k training examples.
	We also test sampling \dataset{} examples uniformly across different tasks and replacing the unified format  (\Cref{sec:apdx_format}) with naive concatenation of tasks and text.
	In the bottom rows, we show the performance of using all \dataset{} training data along with human annotated NLI and classification datasets.
	}\label{tab:diversity_test}
	\resizebox{0.95\linewidth}{!}{%
	\begin{tabular}{l|ccccccc|c}
		\toprule
		 & Class. & Cluster. & Pair. & Rerank. & Retrieval & STS & Summary & Avg. \\
		
		\midrule
		Baseline~\citep{Ni2021LargeDE} &  67.11 & 41.51 & 86.13 & 55.97 & 47.96 & 77.80 & 30.21 & 58.42 \\
		\midrule
		 \multicolumn{9}{l}{\textit{\dataset{} synthetic data ablation}} \\
		 \midrule
		\dataset{}-question-answering & 69.39 & 45.58 & \textbf{84.40} & 56.30 & 49.65 & 78.98 & 31.17 & 60.32 \\
		\dataset{}-search-result & 70.41 & 44.12 & 82.99 & 56.50 & 49.65 & 78.82 & 31.27 & 60.17 \\
		\dataset{}-fact-checking & \textbf{70.81} & \textbf{45.70} & 81.63 & \textbf{57.31} & 49.38 & 79.34 & 30.99 & 60.56 \\
		\dataset{}-sentence-similarity & 70.25 & 45.60 & 81.46 & 56.73 & 47.26 & \textbf{82.02} & \textbf{31.80} & 60.30 \\

		\midrule
        \dataset{}-all-tasks (300K) & 70.25 & 44.56 & 85.37 & 56.46 & \textbf{50.19} & 80.07 & 30.67 & 60.70 \\
        $[+]$ Uniform task sampling & 70.57 & 45.00 & 85.35 & 56.84 & 49.67 & 80.70 & 31.34 & \textbf{60.87} \\
        $[-]$ Unified format & 61.72 & 45.58 & 82.89 & 54.52 & 45.82 & 79.06 & 30.29 & 57.45 \\
        
        \midrule 
        \multicolumn{9}{l}{\textit{Human data ablation}} \\ \midrule
        \dataset{} (6.6M) & 70.26 & 46.82 & 86.27 & 57.60 & 53.16 & 83.14 & 32.16 & 62.64 \\
        $[+]$ NLI datasets & 71.86 & 46.91 & 86.60 & 57.51 & 52.93 & 84.74 & 32.11 & 63.24 \\
        $[+]$ Class. datasets & 81.00 & 46.85 & 86.13 & 57.80 & 52.84 & 82.78 & 32.35 & 64.82 \\
        $[+]$ Full mixture & \textbf{81.17} & \textbf{47.48} & \textbf{87.61} & \textbf{58.91} & \textbf{55.70} & \textbf{85.06} & \textbf{32.63} & \textbf{66.31} \\
        \bottomrule
	\end{tabular}
	}
\end{table}
\subsection{Analysis}
\label{sec:analysis}
\paragraph{LLM as a Labeler}
In \Cref{tab:label_test}, we test different labeling strategies for \dataset{} where we use different positive and hard negative passages.
For positive passages, we try 1) the original passage where the queries were generated (i.e. $p_\text{seed}$), or 2) the top-1 passage selected by an LLM out of the nearest neighbor passages (including the original one) of a generated query (i.e. $p_1$).
For negative passages, we try 1) a random nearest neighbor passage that is different from the original passage (i.e. $p \sim P \setminus \{p_\text{seed}\}$), or 2) the $k$-th passage as ranked by the LLM out of the nearest neighbor passages (including the original one) for the given query (i.e. $p_k$).
From the result, we find that using the most relevant passage chosen by an LLM is always better than using the original passage as positive.
This implies that the original passage is not necessarily best passage to use as a positive target despite the fact that the query was generated from it.
In our qualitative analysis in \Cref{tab:fret_examples}, we show that such cases happen quite often.
\begin{table*}[t]
    \small
    \centering
    \resizebox{0.98\linewidth}{!}{%
    \begin{tabular}{lp{15.0cm}}
        \toprule
               
        \textbf{Seed Passage} ($p_\text{seed}$) & Recently, Marvel’s The Eternals has become the topic of a great deal of online discourse, in part because of a scene where Phastos, a character blessed with the power of invention, helps humanity create the atomic bomb. As you can probably imagine, Twitter saw this and lost it. \\
        \midrule
        Generated Task ($t$) & \textit{Given a query, find a passage that has the answer to the query.} \\
        Generated Query ($q$) & \textit{who made the atomic bomb?}  \\
        LLM-mined Positive ($p_1$) & The film follows the story of American scientist J. Robert Oppenheimer and his role in the development of the atomic bomb.  \\
        LLM-mined Negative ($p_{20}$) & Amid deepening crises around the world with nuclear undertones, a research team from the University of Tokyo will hold a digital exhibition in New York to convey the testimonies of A-bomb survivors on the sidelines of the United Nations review conference of a nuclear nonproliferation treaty.   \\
        
        \midrule
        \vspace{-0.5cm}\\
        \midrule
        \textbf{Seed Passage} ($p_\text{seed}$) &  moose - online shopping for canadians. The 2010 Vancouver Winter Olympics \$75 gold coins were sold individually or in sets of three coins. The three different sets offered were Canadian Wildlife, Canadian Emblems and Vancouver 2010 Olympic Winter Games.    \\
        \midrule
        Generated Task ($t$) & \textit{Given a query, find a passage that might show up as a search result.} \\
        Generated Query ($q$) & \textit{2010 olympic winter games} \\ 
        LLM-mined Positive ($p_1$) &  The 2010 Winter Olympics return to North America on February 12th, when the world of snow sport enthusiasts descend upon one of North America's most beautiful cities, Vancouver. \\
        LLM-mined Negative ($p_{20}$) &  Published: 9:42pm, 12 Feb, 2018  High winds caused havoc at the Pyeongchang Winter Games on Monday as Olympics chief Thomas Bach dismissed concerns North Korea had tried to “hijack” the competition for political gain.   \\
                \midrule

        \vspace{-0.5cm}\\
        \midrule
        \textbf{Seed Passage} ($p_\text{seed}$) & Tagged: Batman, Robin, DC, DC Comics, Comics, ...  \\
        \midrule
        Generated Task ($t$) & \textit{Given a query, find a passage that allows you to check whether the query is true or not.} \\
        Generated Query ($q$) & \textit{Batman is from DC comics} \\
        LLM-mined Positive ($p_1$) & The Batman is an American superhero film based on the DC Comics character of the same name. Produced by DC Films and distributed by Warner Bros. Pictures, it is a reboot of the Batman film franchise.   \\
        LLM-mined Negative ($p_{20}$) & "One of my employees wants to dress up in Batman attire," Gaskins said. "As long as he's at work, I told him it was fine." New York Times News Service contributed to this report. \\

        \bottomrule
    \end{tabular}
    }
    \caption{Examples for LLM-mined positives and negatives.
    While the intent of each query aligns with each task, LLM-mined positive is often more relevant than the seed passage for the generated query.
    }
    
    \label{tab:fret_examples}
\end{table*}

\paragraph{Diversity of \dataset{}}
\dataset{} provides queries in multiple tasks including question answering, search result, fact checking, and sentence similarity.
In \Cref{tab:diversity_test}, we test how the diversity of \dataset{} influences model generalizability across tasks in MTEB.
First, we train individual models each using 300k data from a specific task (e.g., \dataset{}-question-answering).  Additionally, we train models on 300k samples drawn across all four tasks (75k per task; \dataset{}-all-tasks) with original sampling distribution or uniform sampling distribution.  We observe superior performance from the \dataset{}-all-tasks model, particularly when tasks were uniformly sampled.
We also find that the unified formatting (\Cref{sec:apdx_format}) affects the quality of embeddings significantly, as it helps the model better separate different tasks.

\paragraph{Learning Semantic Similarity and Classification}
In the last rows of \Cref{tab:diversity_test}, we show how \ours{} learns better semantic similarity and classification.
We use the symmetric format (Sym.) as well as the same tower negatives for learning better semantic similarity.
Along with the NLI datasets, it drastically improves the STS performance by 1.6 on average.
Our strategy of combining classification datasets also improve the performance on classification by a large margin without significant performance degradation on other tasks.
Using the full \dataset{} mixture gives us the final performance of 66.31.

\paragraph{Qualitative Analysis}
 
  \Cref{tab:fret_examples} showcases the advantages of LLM relabeling. We provide examples of the original seed passage, generated task and query, and the LLM-mined positive and negative passages. First, we observe that the LLM does generate diverse tasks and queries by conditioning on seed passages $p_\text{seed}$. Second, the table highlights the LLM's ability to find a passage ($p_1$) that provides a more direct and relevant answer to the generated query than the seed passage ($p_\text{seed}$). Furthermore, LLM-ranked hard negatives make a challenging task of understanding nuanced differences. 
  These examples demonstrate how the 2-step LLM distillation process effectively brings the LLM's diverse domain knowledge and global ranking preferences into the text embedding model.

\section{Conclusion}

In this paper, we introduced \ours{}, a versatile text embedding model distilled from large language models.
\ours{} is trained on an LLM-generated synthetic dataset \dataset{} that contains LLM-ranked positives and negatives.
We demonstrate that LLMs can be used to identify better positive as well as negative targets for synthesized queries.
We also show how combining this synthetically-generated data in a unified format can lead us to achieve great performance on multiple different tasks at the same time.
Our ablation study reveals the importance of LLM-based relabeling and the diversity of the datasets while demonstrating the strong zero-shot generalizability of \ours{}.

\bibliographystyle{abbrvnat}
\nobibliography*
\bibliography{bibtex}

\clearpage

%

\section*{Author Contributions}
\textbf{Jinhyuk Lee}: Co-lead of FRet and Gecko. Coordinated the project, implemented the main functionality of FRet and Gecko, and led the paper writing.
\textbf{Zhuyun Dai}: Co-lead of FRet. Implemented the main functionality of FRet and led the paper writing.
\textbf{Xiaoqi Ren}: Co-lead of Gecko. Implemented the main functionality of Gecko and its multilingual version.
\textbf{Blair Chen}: Contributed to the MTEB evaluation and ablation study of \ours{}.
\textbf{Daniel Cer}: Contributed to the MTEB evaluation of \ours{} and the classification datasets used for \ours{}.
\textbf{Jeremy R. Cole}: Contributed to experiments for generating and filtering FRet and paper writing.
\textbf{Kai Hui}: Contributed to the use of LLM as a labeler, rank fusion, and paper writing.
\textbf{Michael Boratko}: Contributed to the project coordination and paper writing.
\textbf{Rajvi Kapadia}: Contributed to the use of LLM for the distillation.
\textbf{Wen Ding}: Contributed to the hyperparameter tuning and ablation study of \ours{}.
\textbf{Yi Luan}: Contributed to the use of LLM as a labeler and paper writing.
\textbf{Sai Meher Karthik Duddu}: Contributed to the large-scale training of \ours{}.
\textbf{Gustavo Hernandez Abrego}: Contributed to the project coordination.
\textbf{Weiqiang Shi}: Contributed to the multilingual version of \ours{}.
\textbf{Nithi Gupta}: Contributed to the MRL implementation.
\textbf{Aditya Kusupati}: Contributed to the MRL implementation.
\textbf{Prateek Jain}: Contributed to the MRL implementation.
\textbf{Siddhartha Reddy Jonnalagadda} Contributed to the project coordination.
\textbf{Ming-Wei Chang}: Contributed to the project coordination and paper writing.
\textbf{Iftekhar Naim}: Contributed to the project coordination and paper writing. 

\section*{Acknowledgements}
We thank Devendra Singh Sachan, Michael Kwong, Slav Petrov, and other internal reviewers from Google for reviewing our paper.
We also thank Umangi Jain for the preliminary experiments on MRL.




\newpage
\appendix
\section*{Appendix}
\section{Enhancing Few-shot LLM Ranking with Ensembling} \label{sec:apdx_ensemble}
To validate the quality of the few-shot reranking, we retrieve the top 100 candidate documents and rerank them using our few-shot LLM reranker.
We compare the performance of two LLM rerankers introduced in \cref{sec:fret}: query likelihood (QL) and relevance classification (RC).
Additionally, we investigate the ensemble of these rerankers using Reciprocal Rank Fusion (RRF): $R(q, p) = 1/r_\text{QL}(q,p) + 1/r_\text{RC}(q,p)$, where  $r_\text{QL}(q,p) > 0$  and $r_\text{RC}(q,p) > 0$ represent the rank positions assigned to passage $p$ by QL and RC models for query $q$, respectively.
It is important to note that we employ the identical prompts $\mathbb{P}_\text{QL}$ and $ \mathbb{P}_\text{RC}$ used in \cref{sec:fret}, but not a task-specific prompt for each BEIR task.

\begin{table}[ht]
\small
\centering

\resizebox{0.95\linewidth}{!}{%
\begin{tabular}{l|ccccccccccccc|c}
\toprule
    & CV & NF & TO & DB & SF & CF & HQ & FQ & SD & FE & AR & QU & NQ & \textbf{Avg.} \\
    \midrule
    \multicolumn{14}{l}{\textit{Trained on MS-MARCO}} \\
    \midrule
    RankLLAMA &  85.2 & 30.3 & 40.1 & 48.3 & 73.2 & 28.0 & 75.3 & 46.5 & 17.8 & 83.9 & 56.0 & 85.0 & 66.3 & {56.6} \\
    \midrule
    \multicolumn{14}{l}{\textit{Our Few-shot Prompted LLM Re-Rankers}} \\
    \midrule
    Baseline  & 72.7 & 38.1 & 21.3 & 39.7 & 71.7 & 23.6 & 64.4 & 49.0 & 16.4 & 81.6 & 51.1 & 85.3 & 51.7 &  {51.3} \\ 
    $[+]$ QL &78.8 & 40.9 & 21.3 & 43.8 & 75.2 & \textcolor{red}{15.2} & 76.1 & 57.1 & 22.1 & \textcolor{red}{76.6} & \textcolor{red}{35.7} & 86.3 & 57.3  & {52.8} \\
    $[+]$ RC  & 83.7 & 40.6 & 21.9 & 45.3 & 74.2 & 24.8 & \textcolor{red}{62.3} & \textcolor{red}{46.8} & 20.3 & \textcolor{red}{71.1} & 59.9 & \textcolor{red}{85.0} & 66.9 & {54.1} \\
    $[+]$ RRF(QL, RC) & 84.1 & 41.9 & 22.9 & 46.8 & 76.8 & 22.0 & 76.0 & 56.7 & 22.7 & \textcolor{red}{78.8} & 55.6 & 87.2 &  66.5 & \textbf{56.8} \\

\bottomrule
\end{tabular}
}
\caption{Few-shot LLM re-ranking performance on BEIR. We use the standard nDCG@10 metric. We report results from RankLLAMA~\citep{ma2023rankllama}, a state-of-the-art re-ranker trained on MS-MARCO, for comparison.  \textcolor{red}{Red} indicates that the re-ranker is worse than the baseline retriever. }\label{tab:ensemble_reranker}
\end{table}

\vspace{-0.2cm}
\Cref{tab:ensemble_reranker} shows the results. Reranking with either QL or RC improves the performance. 
Ensembling (RRF) significantly improves the overall quality.  Importantly, the ensembled reranker consistently improves the initial retriever across all tasks except for FEVER (FE), which highlights its robustness to different tasks.
This is important for creating the \dataset{} dataset since we need high quality retrieval data across a diverse range of tasks.

\vspace{-0.3cm}
\section{Formatting in \dataset{}}\label{sec:apdx_format}
Since we aggregate multiple datasets from different tasks, we preprocess every input and target with a unified encoding format.
In \Cref{tab:fret_format}, we show that the performance of asymmetric tasks (i.e. BEIR) is sensitive to the format while the performance of symmetric tasks are relatively stable.
\begin{table*}[ht]
    \small
    \centering
    \resizebox{0.41\linewidth}{!}{%
    \begin{tabular}{l|l}
         \toprule
         \multicolumn{2}{l}{\textit{Symmetric Formatting}} \\
         \midrule
         Input & \texttt{ task: \{task\} | query: \{input\} }\\
         Target & \texttt{ task: \{task\} | query: \{target\} }\\
         
         \midrule
         \multicolumn{2}{l}{\textit{Asymmetric Formatting}} \\
         \midrule
         Input & \texttt{ task: \{task\} | query: \{input\} }\\
         Target & \texttt{ title: \{title\} | text: \{target\} }\\
         
         \bottomrule
    \end{tabular}
    }
    \quad
    \resizebox{0.56\linewidth}{!}{%
    \begin{tabular}{l|cc}
         \toprule
         Formatting & BEIR & STS \\
         \midrule
         Input = \texttt{ \{task\} \{input\} }  & \multirow{2}{*}{54.7} & \multirow{2}{*}{84.8} \\
         Target = \texttt{ \{title\} \{target\} } \\
         
         
         \midrule
         Input = \texttt{ The task is \{task\}, and the query is \{input\} } & \multirow{2}{*}{54.5} & \multirow{2}{*}{\textbf{85.0}} \\
         Target = \texttt{ The title is \{title\}, and the text \{target\} } \\
         
        
        
         \midrule
         Input = \texttt{ task: \{task\} | query: \{input\} } & \multirow{2}{*}{\textbf{55.5}} & \multirow{2}{*}{84.9}\\
         Target = \texttt{ title: \{title\} | text: \{target\} } \\
         \bottomrule
    \end{tabular}
    }
    \caption{Formatting for \dataset{} and other mixture datasets.
    We standardize different datasets and tasks in a unified encoding format (left).
    We also show the performance on BEIR (asymmetric formatting) and STS (symmetric formatting) with different formats (right).
    }
    \label{tab:fret_format}
\end{table*}

\vspace{-0.8cm}
\section{Full MTEB Results and Instructions}\label{sec:full_mteb}
In \Cref{tab:mteb_detail}, we show the full MTEB results. In \Cref{tab:mteb_instruction}, we show the task strings (or instructions) used in the MTEB evaluation. Note that we use consistent instructions for most tasks except for BEIR, which contains multiple different intents as described in \citet{dai2022promptagator}.
\begin{table}[ht]
\footnotesize
    \centering

\resizebox{0.75\linewidth}{!}{%
    \begin{tabular}{l l c c}
\toprule
 & \textbf{Dataset} & \textbf{\ours{}-1B-256} & \textbf{\ours{}-1B-768} \\ \midrule
 \multirow{11}{*}{Classification} 
 & AmazonCounterfactualClassification & 70.93 & 75.34  \\
 & AmazonPolarityClassification & 97.34 & 97.34 \\
 & AmazonReviewsClassification & 48.47 & 51.17 \\
 & Banking77Classification & 86.01 & 88.62 \\
 & EmotionClassification & 51.53 & 52.51\\
 & ImdbClassification & 95.07 & 95.65 \\
 & MTOPDomainClassification & 98.02 & 98.35 \\
 & MTOPIntentClassification & 77.82 & 83.43 \\
 & MassiveIntentClassification & 75.67  & 80. 22 \\
 & MassiveScenarioClassification & 85.16 & 87.19  \\
 & ToxicConversationsClassification & 88.33 & 89.67\\
 & TweetSentimentExtractionClassification& 72.97 & 74.52\\ \midrule
  \multirow{3}{*}{Classification Pair} 
 & SprintDuplicateQuestions & 96.49 & 96.26\\
 & TwitterSemEval2015 & 78.23 & 79.04 \\ 
 & TwitterURLCorpus & 87.04 & 87.53 \\ \midrule
  \multirow{10}{*}{STS} 

 & BIOSSES & 89.42  & 89.46\\
 & SICK-R & 81.67 & 81.92\\
 & STS12 & 78.02 & 77.59 \\
 & STS13 & 90.10 & 90.36 \\
 & STS14 & 85.44 & 85.25\\
 & STS15 & 89.64 & 89.66 \\
 & STS16 & 87.24 & 87.34\\
 & STS17 & 90.46 & 92.06\\
 & STS22 &  67.99 & 68.02 \\
 & STSBenchmark & 89.33 & 88.99\\ \midrule
  \multirow{11}{*}{Clustering} 
 & ArxivClusteringP2P & 44.12 & 46.27 \\
 & ArxivClusteringS2S & 36.54 & 38.36 \\
 & BiorxivClusteringP2P & 36.28 & 37.87 \\
 & BiorxivClusteringS2S & 33.09 & 35.67 \\
 & MedrxivClusteringP2P & 32.08 & 33.11 \\
 & MedrxivClusteringS2S & 30.84 & 31.54 \\
 & RedditClustering & 62.24 & 65.81\\
 & RedditClusteringP2P & 63.70 & 66.62 \\
 & StackExchangeClustering & 70.19 & 74.52\\
 & StackExchangeClusteringP2P & 36.10 & 37.63 \\ 
 & TwentyNewsgroupsClustering & 50.60 & 54.87 \\\midrule
  \multirow{4}{*}{Reranking} 
 & AskUbuntuDupQuestions & 63.84 & 64.40\\
 & MindSmallReranking & 31.89 & 33.07\\
 & SciDocsRR & 81.62 & 83.59\\
 & StackOverflowDupQuestions & 53.76 & 54.56\\\midrule
  \multirow{15}{*}{Retrieval} 
 & ArguAna & 56.27 & 62.18\\
 & ClimateFEVER & 29.35 & 33.21\\
 & CQADupstackAndroidRetrieval & 45.44  & 48.82\\
 & DBPedia & 41.91 & 47.12\\
 & FEVER & 82.61 & 86.96 \\
 & FiQA2018 & 55.54 & 59.24\\
 & HotpotQA & 64.65 & 71.33 \\
 & MSMARCO & 31.12 & 32.58\\
 & NFCorpus & 37.81 & 40.33\\
 & NQ & 57.37 & 61.28\\
 & QuoraRetrieval & 87.89 & 88.18\\
 & SCIDOCS & 18.21 & 20.35 \\
 & SciFact & 70.86 & 75.42\\
 & TRECCOVID & 80.13 & 82.62\\ 
 & Touche2020 & 27.40 & 25.86\\\midrule
  \multirow{1}{*}{Summarization} 
 & SummEval & 32.36 & 32.63\\
 \midrule
 \multicolumn{2}{l}{\textbf{Average}} & 64.37 & 66.31 \\

 \bottomrule
    \end{tabular}}
    \caption{Results for each dataset in the MTEB benchmark.}
    \label{tab:mteb_detail}
\end{table}
\begin{table}[ht]
\small
\centering
\resizebox{0.75\linewidth}{!}{%
\begin{tabular}{llcc}
\toprule
 & \textbf{Dataset} & Task Type & Symmetric \\ \midrule
 \multirow{11}{*}{Classification} 
 & AmazonCounterfactualClassification & classification & \checkmark \\
 & AmazonPolarityClassification & classification & \checkmark \\
 & AmazonReviewsClassification & classification & \checkmark \\
 & Banking77Classification & classification & \checkmark \\
 & EmotionClassification & classification & \checkmark \\
 & ImdbClassification & classification & \checkmark \\
 & MTOPDomainClassification & classification & \checkmark \\
 & MTOPIntentClassification & classification & \checkmark \\
 & MassiveIntentClassification & classification & \checkmark \\
 & MassiveScenarioClassification & classification & \checkmark \\
 & ToxicConversationsClassification & classification & \checkmark \\
 & TweetSentimentExtractionClassification & classification & \checkmark \\ \midrule
  \multirow{3}{*}{Classification Pair} 
 & SprintDuplicateQuestions & semantic similarity & \checkmark \\
 & TwitterSemEval2015 & semantic similarity & \checkmark \\
 & TwitterURLCorpus & semantic similarity & \checkmark \\ \midrule
  \multirow{10}{*}{STS} 

 & BIOSSES & semantic similarity & \checkmark \\
 & SICK-R & semantic similarity & \checkmark \\
 & STS12 & semantic similarity & \checkmark \\
 & STS13 & semantic similarity & \checkmark \\
 & STS14 & semantic similarity & \checkmark \\
 & STS15 & semantic similarity & \checkmark \\
 & STS16 & semantic similarity & \checkmark \\
 & STS17 & semantic similarity & \checkmark \\
 & STS22 & semantic similarity & \checkmark \\
 & STSBenchmark & semantic similarity & \checkmark \\ \midrule
  \multirow{11}{*}{Clustering} 
 & ArxivClusteringP2P & search result & \checkmark \\
 & ArxivClusteringS2S & search result & \checkmark \\
 & BiorxivClusteringP2P & search result & \checkmark \\
 & BiorxivClusteringS2S & search result & \checkmark \\
 & MedrxivClusteringP2P & search result & \checkmark \\
 & MedrxivClusteringS2S & search result & \checkmark \\
 & RedditClustering & search result & \checkmark \\
 & RedditClusteringP2P & search result & \checkmark \\
 & StackExchangeClustering & search result & \checkmark \\
 & StackExchangeClusteringP2P & search result & \checkmark \\
 & TwentyNewsgroupsClustering & search result & \checkmark \\\midrule
  \multirow{4}{*}{Reranking} 
 & AskUbuntuDupQuestions & question answering &  \\
 & MindSmallReranking & semantic similarity & \\
 & SciDocsRR &	question answering &  \\
 & StackOverflowDupQuestions & search result & \\\midrule
  \multirow{15}{*}{Retrieval} 
 & ArguAna &  semantic similarity \\
 & ClimateFEVER & search result \\
 & CQADupstackAndroidRetrieval & question answering \\
 & DBPedia & question answering	\\
 & FEVER & search result \\
 & FiQA2018 & question answering \\
 & HotpotQA & search result	\\
 & MSMARCO & question answering	\\
 & NFCorpus & fact checking	\\
 & NQ & question answering\\
 & QuoraRetrieval &  search result & \checkmark	\\
 & SCIDOCS & question answering	\\
 & SciFact & fact checking	\\
 & TRECCOVID & search result	\\
 & Touche2020 & question answering	\\\midrule
  \multirow{1}{*}{Summarization} 
 & SummEval & search result & \\

 \bottomrule
    \end{tabular}}
    \caption{Instruction used for each dataset in the MTEB benchmark. Here, we denote a simplified task type (e.g., question answering) that summarizes each task generated by \ours{}.}
    \label{tab:mteb_instruction}
\end{table}

\end{document}